\documentclass[preprint,12pt]{elsarticle}
\usepackage{amsmath,amssymb,amsfonts}
\usepackage{ulem}
\usepackage{epsfig}
\usepackage{subfigure}
\usepackage{amsmath}
\usepackage{amssymb}
\usepackage{caption}
\usepackage{stfloats}
\usepackage{epstopdf}
\usepackage{graphicx}
\usepackage{textcomp}
\usepackage{diagbox}
\usepackage{algorithm}
\usepackage{algpseudocode}
\usepackage{multirow}

\usepackage{amssymb}

\journal{Elsevier}

\begin{document}

\begin{frontmatter}



\title{Joint Featurewise Weighting and Lobal Structure Learning for Multi-view Subspace Clustering}

\author[1]{Shi-Xun Lin}
\ead{shixunlin@live.com}
\author[2]{Guo Zhong}
\author[3]{Ting Shu}
\address[1]{School of Mathematics and Statistics, Zhaotong University, Zhaotong 657000, China}
\address[2]{Department of Computer and Information Science, University of Macau, Macau, China}
\address[3]{School of Information Technology and Electrical Engineering, The University of Queensland, Brisbane, QLD 4072, Australia}

\begin{abstract}
Multi-view clustering integrates multiple feature sets, which reveal distinct aspects of the data and provide
complementary information to each other, to improve the clustering performance.
It remains challenging to effectively exploit complementary information across multiple views since the original data often contain noise and are highly redundant.
Moreover, most existing multi-view clustering methods only aim to explore the consistency of all views while ignoring the local structure of each view. However, it is
necessary to take the local structure of each view into consideration, because different views would present different geometric structures while admitting the same cluster structure.
To address the above issues, we propose a novel multi-view subspace clustering method via simultaneously assigning weights for different features and capturing local information of data in view-specific self-representation feature spaces. Especially, a common cluster structure regularization is adopted to guarantee consistency among different views. An efficient algorithm based on an augmented Lagrangian multiplier is also developed to solve the associated optimization problem. Experiments conducted on several benchmark datasets demonstrate
that the proposed method achieves state-of-the-art performance. We provide the Matlab code on https://github.com/Ekin102003/JFLMSC.
\end{abstract}




\end{frontmatter}


\section{Introduction}
Clustering is the organization of unlabeled data into similarity groups~\cite{7093199,ZHONG2020127,WANG202073}. Clustering has been widely used for many fields, including data mining~\cite{8417928,8476242}, pattern recognition~\cite{HU2020,SONG2020}, and machine learning~\cite{YANG201948,LOPEZRUBIO201831}. Clustering methods can be mainly categorized into two groups~\cite{JAIN2010651}: partitioning clustering and hierarchical clustering. In particular, spectral clustering~\cite{von2007tutorial} is a graph-based algorithm for partitioning arbitrarily shaped data structure into disjoint clusters. A number of spectral clustering methods have been proposed, such as Ratio Cut~\cite{159993}, K-way Ratio Cut~\cite{1238361}, Min Cut~\cite{244673},
Normalized Cut (NCut)~\cite{868688}, Spectral Embedded Clustering~\cite{6030950}, and so forth. The clustering performance of all of these methods are largely dependent on the quality of the so called similarity graph, which is learned according to the similarities between the corresponding data points.

The recent works~\cite{6482137,7893729,8259470} of spectral clustering-based subspace clustering have attracted considerable attention due to the promising performance in data clustering. Subspace clustering assumes that the data points are drawn from multiple low-dimensional subspaces, i.e., each subspace is equivalent to a cluster. By exploiting the self-expressiveness property of the data~\cite{6482137,7968309} and imposing a properly chosen constraint on the representation coefficients, a high-quality similarity graph which uncovers the intrinsic
structure of data can be obtained. For example, Elhamifar and Vida~\cite{6482137} proposed a sparse subspace clustering (SSC), which learns a graph via adaptively and flexibly selecting data points. To capture the global structure of data, Liu et al.~\cite{6180173} imposed a low-rank constraint on the representation matrix. Dornaika and Weng~\cite{DORNAIKA2019285} incorporated a manifold regularization term into SSC to capture the manifold structure of data. However, these methods are mostly designed for clustering single-view data.

In the era of big data, many real-world data are represented by multiple distinct feature sets. For example, for images, color information and texture information are two different kinds
of features~\cite{ZHANG202094}. Each kind of feature is a particular view, which often provides compatible and complementary information to each other. Traditional clustering methods can be used to group multi-view data by simply concatenating all views into a monolithic one. However, compatible and complementary information across all views can be typically underutilized. Recently, numerous multi-view clustering approaches~\cite{8928520,Nie2017MultiViewCA,luo2018consistent,shu2019multi,10.5555/3060832.3060884,8502831} have been available. Bickel and Scheffer~\cite{1410262} extended the classic $K$-means and expectation-maximization clustering methods to the multi-view case to deal with text data with two views. However, they are directly conducted in the original feature space, and then they fail to discover the geometrical structure of the multi-view data space. Canonical correlation analysis~\cite{10.1145/1553374.1553391} is a prominent statistical approach for learning from multiple views, but are restricted to linear transformation for each view. Liu et al.~\cite{liu2013multi} proposed a multi-view clustering via joint nonnegative matrix factorization algorithm, which learns a common coefficient matrix of all views and use it to clustering. Such a common coefficient matrix may fail to consider the flexible structures of different views because of heterogeneity among views. Assuming that each sample should share the same cluster in all views, co-regularized spectral clustering (CRMSC)~\cite{10.5555/2986459.2986617} proposed two co-regularization terms to enforce the consistency among different views. The Laplacian matrix used in CRMSC is learned from the original data. As a result, the clustering performance may be seriously affected. To learn a reliable similarity graph from different views, low-rank and sparse decomposition are also used in a robust multi-view spectral clustering method~\cite{10.5555/2892753.2892850}. However, the above methods ignore the importance of different views, i.e., treating all the features equally.

Recently, several graph learning-based multi-view clustering methods~\cite{Nie2017MultiViewCA,10.5555/3060832.3060884,8387526,10.5555/3172077.3172245,HU2020251} have been proposed to identify clustering ability of different views. Multi-view learning with adaptive neighbours~\cite{Nie2017MultiViewCA} can automatically learn the weights of all views during each iteration. Simultaneously, the local manifold structure of multi-view data is captured. Auto-weighted multiple graph learning~\cite{10.5555/3060832.3060884} simultaneously learns an optimal weight for each Laplacian matrix and conducts spectral clustering. The above methods achieve better performances compared to those methods that do not consider the weights of views. However, the similarity graph is usually learned in the original feature space. Such a graph might be severely damaged, and the true similarities among samples cannot be guaranteed due to the influence of noise and redundancy.

The above-mentioned work assumes that data lie on a single subspace, which is contradicted with the observation, i.e., in many problems, data in a class often lie in a low-dimensional subspace of the high dimensional ambient space~\cite{5714408}. Recent studies~\cite{gao2015multi,ZHENG202089,WENG2020375,BRBIC2018247,luo2018consistent,8502831} have done many efforts to develop multi-view subspace clustering methods. For example, Gao et al.~\cite{gao2015multi} unify the self-representation learning and spectral clustering into a framework. Brbi\'{c} and Kopriva~\cite{BRBIC2018247} introduced simultaneous sparsity and low-rankness constraints on the representation matrix to capture the underlying information of multi-view data. Luo et al.~\cite{luo2018consistent} jointly exploited consistency and specificity for subspace representation learning via using a shared consistent representation and a set of specific representations. The above methods mainly treat all features of each view as a whole to learn a single representation or common representation of all views. However, features of real data may be a high degree of redundancy, and hence unrelated information can be introduced to degrade clustering performance. Although a large number of feature selection methods have been proposed, performing feature selection and clustering in two separate steps may not obtain the optimal clustering results~\cite{ZHONG2020105512}.

Despite the recent progress within multi-view clustering, most existing methods ignore the feature-level relationship or the local structure of multi-view data. To address the limitations above, we propose a novel multi-view subspace clustering method based on data self-representation, which simultaneously weights features and learns the local structure of each view. By weighting the original features, we can extract the effective and robust features, thus alleviating the influence of redundancy. Moreover, data self-representation and local structure learning are integrated into a unified framework such that the inherent difference in each view can be captured via simultaneously exploiting the global and local information of each view. In particular, a common cluster structure regularization is used to capture consistency across different views.

The major contributions of this paper are listed as follows.

\begin{enumerate}
\item We propose a novel multi-view subspace clustering approach that captures view-specific information from each independent view via simultaneously exploiting the underlying global representation information and local distance information from each view while taking consistency of all views into consideration by spectral embedding.

\item We learn view-specific weight vectors for each type of feature so that the influence of redundancy and noise can be alleviated, and the similarities between samples can be accurately obtained according to a distance regularization term.

\item An augmented Lagrangian alternating direction minimization method is developed to solve the optimization problem. Experiments on diverse multi-view datasets show the effectiveness of the proposed method.
\end{enumerate}

We organize the remainder of this paper as follows. In Section II, we review relevant work. Section III presents a detailed derivation of the proposed multi-view
subspace clustering approach.  Section IV provides an optimization algorithm for solving the minimization problem. We verify our proposed method through extensive experiments in Section V. Finally, section VI concludes this paper.

\section{RELATED WORK}
In this paper, we denote the input sample matrix by $X$ and utilize the normal italic uppercase letters to denote matrices. Symbols are summarized in Table~\ref{table:simbol}.
In this section, we review the most relevant clustering work, i.e., spectral clustering and spectral-clustering-based multi-view subspace clustering methods.
\begin{center}
\begin{table*}
\caption{Description of the Symbols}
\label{table:simbol}
\setlength{\tabcolsep}{60pt}
\begin{tabular}{|p{35pt}|p{135pt}|}
\hline
Symbols&
Description\\
\hline
$n$&
number of samples\\
$c$&
number of clusters\\
$n_v$&
number of views\\
$d_v$&
dimension of views $v$\\
$I$&
unit matrix\\
$X\geq 0$&
nonnegative matrix\\
$x_i$&
$i$-th column of a matrix $X$\\
$x^j$&
$j$-th row of $X$\\
$x_{ij}$&
$ij$-th entry of $X$\\
$\mathbf{1}$&
column vector of ones\\
$\mathbf{0}$&
column vector of zeros\\
$\|X\|_F$    & $\|X\|_F = \sqrt{\sum_{i,j}x_{ij}^2}$\\
$X^{(v)}$  &   data matrix of the $v$-th view\\
$\|X\|_1$    & $\|X\|_1 = \sum_{i,j}|x_{ij}|$\\
$\|x\|_2$    & 2-norm of a vector $x$\\
$X^T$  &   transpose of $X$\\
$Tr(X)$  &   trace of $X$\\
$diag(u)$  &   square diagonal matrix with the elements of vector $u$ on the main diagonal\\
$diag(X)$  &   column vector of the main diagonal elements of $X$\\
\hline
\end{tabular}
\label{tab1}
\end{table*}
\end{center}
\subsection{Spectral clustering}
Assume that a given data matrix $X=[x_1,\ldots,x_n]\in \mathbb{R}^{d\times n},$  we can represent the relationships
between samples in the form of a similarity graph $G = (V,E)$ where $V$ is a set of vertices and $E$ is a set of edges. Each vertex of $G$ represents a sample, and each
edge has a weight reflecting the similarity between two connected samples. The goal of spectral clustering is to partition the similarity graph such that the edges between different groups have very low weights, and the edges within a group have high weights~\cite{von2007tutorial}. To this end, the objective of Ratio Cut spectral clustering method can be formulated as
\begin{equation}\min_{Q^TQ=I,Q\in \mathbb{R}^{n\times c}}\quad Tr(Q^TLQ),\label{01}\end{equation}
where the optimal solution $Q^*$ of this problem consists of the first $c$ eigenvectors of $L.$ $L=D-S$ is the Laplacian matrix, where $S\in \mathbb{R}^{n\times n}$ is the similarity matrix and the diagonal elements of $D$ can be computed by $d_{ii} = \sum_{j=1}^ns_{ij},$ i.e., $D=diag(d_{11},\ldots,d_{nn}).$ The spectral embedding matrix $Q^*$ is regarded as a clustering assignment.

In order to more effectively group multi-view data, multi-view spectral clustering methods~\cite{cai2013heterogeneous,10.5555/3104482.3104532,10.5555/2986459.2986617,10.5555/3060832.3060884,HU2020251} have been proposed. Most of them learn a commonly shared graph Laplacian by unifying different views while neglecting to learn high-quality similarity graphs to improve clustering performance further.

\subsection{Multi-view subspace clustering}

Assume that $X$ is a collection of samples drawn from a union of $c$ linear subspaces $\mathcal{S}_1\cup \mathcal{S}_2\cup\ldots\cup\mathcal{S}_c.$ Subspace clustering groups $x_1,\ldots,x_n$ according to their subspaces.

Spectral clustering-based subspace clustering has demonstrated promising results. In general, its model can be formulated as
\begin{equation}\min_{Z\in \mathbb{R}^{n\times n}}\quad \mathcal{R}(X,XZ)+\lambda\Psi(Z),\label{02}\end{equation}
where $Z=[z_1, z_2,\ldots, z_n]$ is the self-representation matrix. Each column $z_i$ of $Z$ is considered as a new representation of the sample $x_i$ in terms of other samples in $X.$ Those nonzero elements in $Z$ denote that the corresponding samples are from the same cluster. $\lambda> 0$ is the trade-off parameter that balances the regularization term $\Psi(\cdot)$ and the reconstruction error term $\mathcal{R}(\cdot,\cdot).$ Once obtaining the representation matrix $Z$, we can build the similarity graph $S= \frac{|Z|+|Z^T|}{2}.$ Finally, we feed $S$ into a spectral clustering algorithm, such as \cite{1238361}\footnote{The code is available at\: http://www.cis.upenn.edu/~jshi/software/}, to get clustering result.

Recently multi-view subspace clustering methods have been designed to fuse complementary information from multiple views. Cao et al.~\cite{7298657} proposed to utilize the Hilbert Schmidt Independence Criterion as a diversity term to explore the complementarity of multi-view representations. Wang et al.~\cite{wang2017exclusivity} explored the complementary information between different representations by introducing a novel position-aware exclusivity term. The state-of-the-art consistent and specific multi-view subspace clustering~\cite{luo2018consistent} simultaneously learns a view-consistent representation and a set of view-specific representations for multi-view subspace clustering. The general formulation of multi-view subspace clustering can be written as the following form
\begin{equation}\min_{Z^{(v)}\in \mathbb{R}^{n\times n}} \sum_{v=1}^{n_v}[\mathcal{R}(X^{(v)},X^{(v)}Z^{(v)})+\lambda\Psi(Z^{(v)})].\label{03}\end{equation}
By obtaining optimal $Z^{(v)},v = 1,\ldots,n_v,$ we can output the final clustering results by performing the spectral clustering algorithm\cite{1238361} on the similarity matrix \begin{equation}S=\frac{1}{n_v}\sum_{v=1}^{n_v}\frac{|Z^{(v)}| + |Z^{(v)T}|}{2}.\label{04}\end{equation}
\section{Methodology}
\label{}
In this section, to explore better representation capacity of data and more flexible local manifold structures of different views, we derives the proposed method, which unifies the featurewise weight learning and adaptive local structure learning into a framework.

Previous works~\cite{6418020,8620311,7172559} demonstrate the importance of local structure learning in subspace clustering. Most of them mainly utilized a pre-computed graph regularizer to capture local manifold structure of data, i.e.,
\begin{align}\min_{Z} \frac{1}{2}\sum_{i,j}^n \|z_i-z_j\|_2^2b_{ij} = Tr(Z^TL_BZ),\label{05}\end{align}
where $L_B$ is a Laplacian matrix and $B$ is a pre-computed similarity matrix. By integrating Eq.~\eqref{05} into the self-representation learning framework Eq.~\eqref{02}, the learned representations can preserve the local geometry structure embedded in a high-dimensional ambient space. However, $b_{ij}$ has to be pre-computed. That is, to capture the local structure of data in self-representation learning framework, a two-stage process is adopted, which can result in information loss. We will integrate self-representation learning and the local structure learning into a unified framework. On the other hand, the non-negativity is more consistent with the biological modeling of visual data, and often leads to better performance for data representation~\cite{lin2017low}. Moreover, each view should share a consensus clustering assignment. As a result, by Eq.~\eqref{01} we arrive at the following objective:
\begin{align}
&\min_{\substack{Z^{(v)},E^{(v)},Q}}\quad\sum_{v=1}^{n_v}\sum_{\substack{i=1}}^n\sum_{j=1}^n\|x_i^{(v)}-x_j^{(v)}\|_2^2z_{ij}^{(v)}\label{06}\\
&+\sum_{v=1}^{n_v}[\lambda_2\|E^{(v)}\|_1+\lambda_1\lambda Tr(Q^TL_{Z^{(v)}}Q)],\nonumber\\
&\begin{array}{r@{\quad}r@{}l@{\quad}l}
s.t. \quad X^{(v)} = X^{(v)}Z^{(v)}+E^{(v)},Z^{(v)}\geq 0,Z^{(v)}\mathbf{1}=\mathbf{1},\\
diag(Z^{(v)}) = \mathbf{0},Q^TQ=I,\nonumber\\
\end{array}
\end{align}
where $Q\in \mathbb{R}^{n\times c}$ is the consensus clustering assignment across all views. $L_{Z^{(v)}} = D^{(v)}-\frac{Z+Z^T}{2}$ is the Laplacian matrix, where $D^{(v)}=diag(\frac{Z^{(v)}+Z^{(v)T}}{2}\mathbf{1}).$ $\lambda$ is a positive penalty parameter. Due to the nonnegative constraint, each entry of $Z^{(v)}$ can directly reflect the similarity between $x_i^{(v)}$ and $x_j^{(v)}.$ Concretely, the first term of Eq.~\eqref{06} guides the representation matrix to capture the local structure of data. In practical applications, data may be contaminated by noise. Thus we introduce a sparse error term $\|E^{(v)}\|_1$. Close samples should have large similarity while distant samples should have small or even zero similarity. Specifically, we further constraint $Z^{(v)}\mathbf{1}=\mathbf{1}$ from the point of view of probability, i.e., $0\leq z_{ij}<1.$

To capture the global structure of data, following~\cite{TANG201666}, we will add a 2-norm constraint on the representation matrix instead of low-rank constraint since we may obtain bad representations when the dimension of a certain view is far smaller than the number of samples~\cite{gao2015multi}.
Then we have the following minimization problem:
\begin{align}
\min_{\substack{Z^{(v)},Q,\\ E^{(v)}}}&\sum_{v=1}^{n_v}[\sum_{\substack{i,j}}^n\|x_i^{(v)}-x_j^{(v)}\|_2^2z_{ij}^{(v)} + \lambda_1 Tr(Q^TL_{Z^{(v)}}Q)]\nonumber\\
&+\sum_{v=1}^{n_v}[\lambda_2\|Z^{(v)}\|_2+\lambda_3\|E^{(v)}\|_1],\label{07}\\
&\begin{array}{r@{\quad}r@{}l@{\quad}l}
s.t. \quad X^{(v)} = X^{(v)}Z^{(v)}+E^{(v)},Z^{(v)}\geq 0,\\
diag(Z^{(v)}) = \mathbf{0},Q^TQ=I,Z^{(v)}\mathbf{1}=\mathbf{1},\nonumber\\
\end{array}
\end{align}
where $\|Z^{(v)}\|_2$ is the maximum singular value of $Z^{(v)}.$ $\lambda_1,\lambda_2,$ and $\lambda_3$ are a positive penalty parameter.

Note that the first term of Eq.~\eqref{07} use Euclidean distance computed on the original data to guide the local structure learning. Doing this may not reflect the true closeness between $x_i$ and $x_j$ due to the sensitivity of Euclidean distance to noise and redundancy. To alleviate this problem, we learn the importance of different features. Consequently, our final multi-view subspace clustering model is stated as follows:
\begin{align}
&\min_{\substack{Z^{(v)},Q,\\ E^{(v)},w^{(v)}}}\quad\sum_{v=1}^{n_v}\sum_{i=1}^n\sum_{\substack{j=1}}^n\|W^{(v)}x_i^{(v)}-W^{(v)}x_j^{(v)}\|_2^2z_{ij}^{(v)}\label{08}\\
&+\sum_{v=1}^{n_v}[\lambda_2\|Z^{(v)}\|_2+\lambda_3\|E^{(v)}\|_1 + 2\lambda_1 Tr(Q^TL_{Z^{(v)}}Q)],\nonumber\\
&\begin{array}{r@{\quad}r@{}l@{\quad}l}
s.t. \quad X^{(v)} = X^{(v)}Z^{(v)}+E^{(v)},Z^{(v)}\geq 0,\\
diag(Z^{(v)}) = \mathbf{0},Q^TQ=I,Z^{(v)}\mathbf{1}=\mathbf{1},\nonumber\\
W^{(v)}=diag(w^{(v)}),w^{(v)}\geq0,w^{(v)}\mathbf{1}=1,
\end{array}
\end{align}
where $w^{(v)}\in \mathbb{R}^{1\times d_v}$ is the self-weighted vector for feature of the $v$-th view. $w^{(v)}$ adaptively learns the importance of different features in the $v$-th view.

We can see that the objective function in Eq.~\eqref{08} simultaneously takes into consideration of the global and local structure of each view. Specifically, the local structure is learned through more discriminative features instead of pre-computing a locality constraint on the original feature space. Furthermore, we capture the consistency interweaving in different views according to the clustering assignment matrix $Q$ shared by all views since each view should have a different representation matrix resulted in by the heterogeneity of features from different views, but they have the same clustering structure. We refer to the proposed method as Joint Featurewise Weighting and Lobal Structure Learning for Multi-view Subspace Clustering (JFLMSC).

\section{Optimization algorithm}
We use the augmented Lagrange multiplier (ALM) method to efficiently solve the challenging problem~\eqref{08}. Specifically, the problem~\eqref{08} can be optimized alteratively. We introduce auxiliary variables, i.e., $Z^{(v)}=A^{(v)}$ and $Z^{(v)}=U^{(v)},v\in\{1,\ldots,n_v\}.$ Problem~\eqref{08} is converted to the following optimization problem as:
\begin{align}
&\min_{\substack{Z^{(v)},E^{(v)},A^{(v)},\\ U^{(v)},Q,w^{(v)}}}\sum_{v=1}^{n_v}\sum_{i=1}^n\sum_{\substack{j=1}}^n\|W^{(v)}x_i^{(v)}-W^{(v)}x_j^{(v)}\|_2^2a_{ij}^{(v)}\label{09}\\
& + \sum_{v=1}^{n_v}[\lambda_2\|U^{(v)}\|_2 + \lambda_3\|E^{(v)}\|_1+2\lambda_1Tr(Q^TL_{A^{(v)}}Q)],\nonumber\\
&\begin{array}{r@{}l}
s.t. \quad &X^{(v)} = X^{(v)}Z^{(v)} +E^{(v)},Z^{(v)}=U^{(v)},Z^{(v)}=A^{(v)},\\
&A^{(v)}\geq 0,diag(A^{(v)}) = \mathbf{0},A^{(v)}\mathbf{1}=\mathbf{1},Q^TQ=I,\\
&W^{(v)}=diag(w^{(v)}),w^{(v)}\geq0,w^{(v)}\mathbf{1}=1.\nonumber
\end{array}
\end{align}
The augmented Lagrangian function of the objective function in Eq.~\eqref{09} is
\begin{align}
&\mathcal{L}(Z^{(v)},E^{(v)},Q,A^{(v)},U^{(v)},w^{(v)})\label{10}\\
&=\sum_{v=1}^{n_v}\sum_{i=1}^n\sum_{\substack{j=1}}^n\|W^{(v)}x_i^{(v)}-W^{(v)}x_j^{(v)}\|_2^2a_{ij}^{(v)}\nonumber\\
& + \sum_{v=1}^{n_v}[\lambda_2\|U^{(v)}\|_2 + \lambda_3\|E^{(v)}\|_1+2\lambda_1Tr(Q^TL_{A^{(v)}}Q)]\nonumber\\
&+\sum_{v=1}^{n_v}\left<\Lambda_1^{(v)},X^{(v)}-X^{(v)}Z^{(v)}-E^{(v)}\right>\nonumber\\
&+\sum_{v=1}^{n_v}(\left<\Lambda_2^{(v)},Z^{(v)}-U^{(v)}\right>+\left<\Lambda_3^{(v)},Z^{(v)}-A^{(v)}\right>)\nonumber\\
&+\sum_{v=1}^{n_v}\frac{\mu}{2}\|X^{(v)}-X^{(v)}Z^{(v)}-E^{(v)}\|_F^2\nonumber\\
&+\sum_{v=1}^{n_v}\frac{\mu}{2}(\|Z^{(v)}-U^{(v)}\|_F^2+\|Z^{(v)}-A^{(v)}\|_F^2),\nonumber
\end{align}
where $\Lambda_1^{(v)},\Lambda_2^{(v)}$, and $\Lambda_3^{(v)}$ are Lagrangian multipliers, $\mu$ is a positive penalty scalar, and $\left<\cdot,\cdot\right>$ denotes
the matrix inner product. Now, we can solve Eq.~\eqref{10}  with respect to one variable while fixing the other variables. As we will see, Eq.~\eqref{10} is reduced to six subproblems. The optimization for each subproblem is as follows:

\subsubsection{$Z^{(v)}$-Subproblem:}
While keeping the other variables fixed except $Z^{(v)}$, Eq.~\eqref{10} becomes the following:
\begin{equation}
\begin{split}
\mathcal{L}(Z^{(v)}) &= \|X^{(v)}-X^{(v)}Z^{(v)}-E^{(v)}+\frac{\Lambda_1^{(v)}}{\mu}\|_F^2\\
&+\|Z^{(v)}-U^{(v)}+\frac{\Lambda_2^{(v)}}{\mu}\|_F^2\\
&+\|Z^{(v)}-A^{(v)}+\frac{\Lambda_3^{(v)}}{\mu}\|_F^2.
\end{split}
\end{equation}
It suffices to take the derivative of $\mathcal{L}(Z^{(v)})$ with respect to $Z^{(v)}$ and set it to zero. We obtain the following closed-form solution:
\begin{equation}
Z^{(v)} = (X^{(v)T}X^{(v)}+2I)^{-1}(X^{(v)T}V_1+V_2+V_3),\label{11}
\end{equation}
where $V_1=X^{(v)}-E^{(v)}+\frac{\Lambda_1^{(v)}}{\mu},V_2=U^{(v)}-\frac{\Lambda_2^{(v)}}{\mu}$, and $V_3 = A^{(v)}-\frac{\Lambda_3^{(v)}}{\mu}.$

\subsubsection{$A^{(v)}$-Subproblem:}
By ignoring the irrelevant terms of Eq.~\eqref{10}, $A^{(v)}$ can be obtained by solving the following problem:
\begin{align}
\min_{\substack{A^{(v)}}}&\sum_{i=1}^n\sum_{\substack{j=1}}^n\|W^{(v)}x_i^{(v)}-W^{(v)}x_j^{(v)}\|_2^2a_{ij}^{(v)}\label{12}\\
& + 2\lambda_1Tr(Q^TL_{A^{(v)}}Q)+\frac{\mu}{2}\|Z^{(v)}-A^{(v)}+\frac{\Lambda_3^{(v)}}{\mu}\|_F^2,\nonumber\\
&\begin{array}{r@{}l}
s.t. \quad A^{(v)}\geq 0,diag(A^{(v)}) = \mathbf{0},A^{(v)}\mathbf{1}=\mathbf{1}.\nonumber
\end{array}
\end{align}
Note that
\begin{equation}\frac{1}{2}\sum\limits_{i,j=1}^n\|q^i-q^j\|_2^2a_{ij}^{(v)}=Tr(Q^TL_{A^{(v)}}Q).\label{21}\end{equation}
To simplify the notations, we ignore the view index tentatively. The objective function in Eq.~\eqref{12} can be directly decoupled into rows, enabling us to deal with the following problem individually for every $i$:
\begin{align}
\min_{\substack{a^i}}&\sum\limits_{j=1}^n(\|Wx_i-Wx_j\|_2^2a_{ij} + \lambda_1\|q^i-q^j\|_2^2a_{ij})\label{13}\\
&+\sum\limits_{j=1}^n(\frac{\mu}{2}a_{ij}^2-\mu a_{ij}h_{ij}),\nonumber\\
&\begin{array}{r@{}l}
s.t. \quad a^i\geq 0, a_{ii} = 0,a^i\mathbf{1}=1,\nonumber
\end{array}
\end{align}
where $h_{ij}$ is the $(i,j)$-th element of $H = Z+\frac{\Lambda_3}{\mu}.$ Let \begin{equation}d_{ij}= \|Wx_i-Wx_j\|_2^2+\lambda_1\|q^i-q^j\|_2^2-\mu h_{ij}\end{equation} be the $j$-th element of $d_i\in R^{1\times n}.$ With some algebra, we can state Eq.~\eqref{13} as
\begin{align}
\min_{\substack{a^i\geq 0,a_{ii} = 0,a^i\mathbf{1}=1}}\|a^i+\frac{d_i}{\mu}\|_2^2.\label{14}
\end{align}
The Lagrangian function of Eq.~\eqref{14} is
\begin{equation}\mathcal{L}(a^i,\eta,\beta) = \|a^i+\frac{d_i}{\mu}\|_2^2 + \eta(1-a^i\mathbf{1})+\beta^T(-a^i),\end{equation}
where $\eta$ and $\beta\geq \mathbf{0}$ are the Lagrangian multipliers.
It follows the KKT conditions \cite{boyd2004convex} that the optimal solution $a^i$ is
\begin{align}
a^i=(-\frac{d_i}{\mu}+\eta\mathbf{1}^T)_+,\label{15}
\end{align}
where $(\cdot)_+ = \max(\cdot, 0).$

\subsubsection{$Q$-Subproblem:}
By fixing the other variables to constants, we can obtain $Q$ by solving the following problem:
\begin{align}
\min_{\substack{Q\in \mathbb{R}^{n\times c},Q^TQ=I}} Tr(Q^T(\sum_{v=1}^{n_v} L_{A^{(v)}})Q).\label{16}
\end{align}
The optimal solution $Q^*$ consists of the $c$ eigenvectors of $\sum_{v=1}^{n_v} L_{A^{(v)}}$ with respect to the $c$ smallest eigenvalues.

\subsubsection{$U^{(v)}$-Subproblem:}
To update $U^{(v)}$ with other variables fixed, we solve the following problem
\begin{align}
&\min_{\substack{U^{(v)}}} \lambda_2\|U^{(v)}\|_2 + \frac{\mu}{2}\|Z^{(v)}-U^{(v)}+\frac{\Lambda_2^{(v)}}{\mu}\|_F^2,\label{17}
\end{align}
which can be solved by Proposition 4~\cite{TANG201666}.

\subsubsection{$E^{(v)}$-Subproblem:}
We fix all variables except $E^{(v)}$ to solve the following problem
\begin{align}
&\min_{\substack{E^{(v)}}} \lambda_3\|E^{(v)}\|_1 + \frac{\mu}{2}\|X^{(v)}-X^{(v)}Z^{(v)}-E^{(v)}+\frac{\Lambda_1^{(v)}}{\mu}\|_F^2.\label{18}
\end{align}
Following \cite{10.1145/1970392.1970395}, we can obtain
\begin{equation}
\label{19}
E^{(v)}=\Omega_{\frac{\lambda_3}{\mu}}(X^{(v)}-X^{(v)}Z^{(v)} +\frac{\Lambda_1^{(v)}}{\mu}),
\end{equation}
where $\Omega$ denotes the shrinkage operator.

\subsubsection{$w^{(v)}$-Subproblem:}
When fixing the other variables, the problem~\eqref{10} is equivalent to the following problem:
\begin{align}
\min_{\substack{w^{(v)}}}&\sum_{\substack{i,j}}\|W^{(v)}x_i^{(v)}-W^{(v)}x_j^{(v)}\|_2^2a_{ij}^{(v)}\label{20}\\
&\begin{array}{r@{}l}
s.t. \quad W^{(v)}=diag(w^{(v)}),w^{(v)}\geq 0, w^{(v)}\mathbf{1}=1.\nonumber
\end{array}
\end{align}
According to the Eq.~\eqref{21}, we rewrite problem~\eqref{20} as follows:
\begin{align}
\min_{\substack{w^{(v)}}}&\quad Tr(W^{(v)}X^{(v)}L_{A^{(v)}}X^{(v)T}W^{(v)})\label{22}\\
&\begin{array}{r@{}l}
s.t. \quad W^{(v)}=diag(w^{(v)}),w^{(v)}\geq 0, w^{(v)}\mathbf{1}=1.\nonumber
\end{array}
\end{align}
The problem~\eqref{22} can further transform into the following:
\begin{align}
\min_{\substack{w^{(v)}}}&\quad Tr(W^{(v)}YW^{(v)})\label{23}\\
&\begin{array}{r@{}l}
s.t. \quad W^{(v)}=diag(w^{(v)}),w^{(v)}\geq 0, w^{(v)}\mathbf{1}=1,\nonumber
\end{array}
\end{align}
where $Y= X^{(v)}L_{A^{(v)}}X^{(v)T}.$ We simplify Eq.~\eqref{23} as
\begin{align}
\min_{\substack{w^{(v)}}}&\quad w^{(v)}\mathbf{y}w^{(v)T}\label{24}\\
&\begin{array}{r@{}l}
s.t. \quad w^{(v)}\geq 0, w^{(v)}\mathbf{1}=1,\nonumber
\end{array}
\end{align}
The Lagrangian function of Eq.~\eqref{24} is
\begin{equation}\mathcal{L}(w^{(v)},\eta) = w^{(v)}\mathbf{y}w^{(v)T}- \eta(w^{(v)}\mathbf{1} - 1),\label{25}\end{equation}
where $\mathbf{y}=diag(y_{11},\ldots,y_{d_vd_v})$ and $\eta$ is the Lagrangian multiplier.

By taking the derivative of Eq.~\eqref{25} with respect to $w^{(v)}$ and setting it to zero, we obtain the closed-form solution as follows:
\begin{equation} w^{(v)}=(\frac{1}{y_{11}\sum_{j=1}^{d_v}\frac{1}{y_{jj}}},\ldots,\frac{1}{y_{d_vd_v}\sum_{j=1}^{d_v}\frac{1}{y_{jj}}}),\label{26}\end{equation}
where we use the constraint $w^{(v)}\mathbf{1}=1.$
\subsubsection{Multiplier:}
We update $\Lambda_1^{(v)},\Lambda_2^{(v)},\Lambda_3^{(v)}$, and $\mu$ as follows:
\begin{equation}
\label{27}
\begin{split}
\Lambda_1^{(v)}&=\Lambda_1^{(v)} +\mu(X^{(v)}-X^{(v)}Z^{(v)}-E^{(v)}),\\
\Lambda_2^{(v)}&= \Lambda_2^{(v)} +\mu(Z^{(v)}-U^{(v)}),\\
\Lambda_3^{(v)}& = \Lambda_3^{(v)} + \mu(Z^{(v)}-S^{(v)}),\\
\mu &= \min (\rho\mu,\mu_{max}),
\end{split}
\end{equation}
where $\rho$ and $\mu_{max}$ are prefixed constants, and $\rho$ controls the convergence speed.

We have derived an augmented Lagrange multiplier with alternating direction minimizing (ALM-ADM) algorithm for our proposed JFLMSC. For clarity, we summarize the optimization algorithm to solve problem~\eqref{08} in Algorithm~\ref{algo}. The computational complexity is comparable with many state-of-the-art multi-view subspace clustering methods~\cite{BRBIC2018247,luo2018consistent,8502831}. Theoretically proving the convergence of Algorithm~\ref{algo} is difficult~\cite{ZHENG202089,WENG2020375,BRBIC2018247,luo2018consistent,8502831}. Similar to the previous work~\cite{WENG2020375,BRBIC2018247,luo2018consistent}, we show the convergence empirically in the experiment. However, under mild conditions, any limit point of the iteration sequence generated by Algorithm~\ref{algo} is a stationary point that satisfies the KKT conditions. Detailed convergence analysis can be found in \cite{8740912}.
\begin{algorithm}[t]
  \caption{Solving Problem~\eqref{08} via ALM-ADM Algorithm}
      \label{algo}
  \begin{algorithmic}[1]
    \Require Multi-view data: $X^{(1)}\in \mathbb{R}^{d_1\times n},\ldots,X^{(n_v)}\in \mathbb{R}^{d_{n_v}\times n},$ and regularization parameters $\lambda_1,\lambda_2$ and $\lambda_3.$
    \Ensure
      $Q,Z^{(v)},U^{(v)},E^{(v)},A^{(v)},w^{(v)},\forall v=1,\ldots,n_v.$
    \State initialization: $\mu>0,\rho>1,\Lambda_1^{(v)}=\mathbf{0},\Lambda_2^{(v)}=\Lambda_3^{(v)}=\mathbf{0},Z^{(v)}=A^{(v)},U^{(v)}=Z^{(v)},E^{(v)}=\mathbf{0}$ and $Z^{(v)}$ is an affinity matrix based on $k$ nearest neighbor graph.
    \Repeat
    \While {$v\leq n_v$}
      \State Update $Z^{(v)}$ using Eq.~\eqref{11}.
      \State Update $A^{(v)}$ line by line using Eq~\eqref{15}.
      \State Update $U^{(v)}$ by solving problem~\eqref{17}.
      \State Update $E^{(v)}$ using Eq.~\eqref{19}.
      \State Update $w^{(v)}$ using Eq.~\eqref{26}.
      \State Update $\Lambda_1^{(v)},\Lambda_2^{(v)},$ and $\Lambda_3^{(v)}$ using Eq.~\eqref{27}.
      \EndWhile
      \State Update $Q$ by solving problem~\eqref{16}.
      \State Update $\mu$ using Eq.~\eqref{27}.
    \Until convergence
  \end{algorithmic}
\end{algorithm}

\section{Experiments}
\label{}
In this section, we conduct experiments to evaluate the clustering performance of the proposed methods on four public
data sets. Benchmarking clustering is generally difficult~\cite{luo2018consistent}. Following previous work~\cite{HU2020251}, five evaluation metrics, i.e., ACC (Clustering Accuracy), NMI (Normalized Mutual Information), ARI (Adjusted Rand Index), Precision, and F-score, are adopted.

\subsection{Data sets}

In our experiment, we collected a diversity of four datasets, i.e., 100leaves, MSRC-v1, ORL, and Outdoor Scene, to demonstrate the effectiveness of our proposed JFLMSC.
\begin{enumerate}
\item One-hundred plant species leaves data set (100leaves)~\cite{8662703} consists of 1600 samples from each of one hundred
plant species. For each sample, texture histogram, fine-scale margin, and shape descriptor are given.

\item MSRC-v1 data set~\cite{winn2005object} includes 210 images. Following~\cite{10.5555/2540128.2540503}, we extract GIST with dimension
512, HOG with dimension 100, LBP with dimension 256, SIFT with dimension 210,  color moment with dimension 48, and CENTRIST with dimension 1302 visual features from each image.

\item ORL data set~\cite{samaria1994parameterisation} contains 10 different images of each of 40 distinct subjects. For ORL, we extract three types of features, i.e., the intensity with dimension 4096, LBP with dimension 3304, and Gabor with dimension 6750.

\item Outdoor Scene~\cite{HU2020251} has 2688 images consisting of 8 groups. For each image, we extract GIST with dimension 512, color moment with dimension 432, HOG with dimension 256, and LBP with dimension 48.
\end{enumerate}

We summarized all the data sets in Table~\ref{table:sta}, where d$_v$ denotes the dimension of features in view $v$.

\begin{table}[t]
\begin{center}
\caption{Descriptions of data sets.}
\label{table:sta}
\scalebox{0.9}{
\begin{tabular}{ccccc}
\hline\noalign{\smallskip}
Datasets & 100leaves & MSRC-v1 & ORL & Outdoor Scene\\
\hline\noalign{\smallskip}
Number of samples & 1,600  & 210 & 400 & 2,688\\
\hline\noalign{\smallskip}
Number of clusters &  100 & 7 & 40 & 8\\
\noalign{\smallskip}
\hline
\noalign{\smallskip}
\#d$_1$  & 64  & 254 & 4,096& 432\\
\#d$_2$ &  64 &  24& 3,304 & 256\\
\#d$_3$ & 64 & 512& 6,750 & 512\\
\#d$_4$ &  - & 576& -& 48\\
\#d$_5$ & - & 256& -& -\\
\hline
\end{tabular}}
\end{center}
\end{table}

\subsection{Compared methods}
We compare the proposed method with the following baseline methods, including several closely related state-of-the-art methods.

\begin{enumerate}
\item Normalized Cut (NCut)~\cite{868688} is a traditional spectral clustering approach for clustering single view data. The features of each view of multi-view data are concatenated in a column-wise to feed into NCut.

\item Co-regularized Multi-view Spectral Clustering (CRMSC) \cite{10.5555/2986459.2986617} is a spectral clustering framework that achieves exploiting information from multiple views by co-regularizing the consistency across the views.

\item Multi-View Clustering with Adaptive Neighbours (MCAN)~\cite{Nie2017MultiViewCA} is a novel multi-view learning model that performs clustering and local structure learning simultaneously. Moreover, MCAN can allocate an ideal weight for each view automatically.

\item Auto-Weighted Multiple Graph Learning (AMGL)~\cite{10.5555/3060832.3060884} is a novel framework via the reformulation of the standard spectral learning model, which
can learn an optimal weight for each graph automatically.

\item Self-weighted Multi-view Clustering with Multiple Graphs (SWMC)~\cite{10.5555/3172077.3172245} explores a Laplacian rank constrained graph, which can be approximate as the centroid of the built graph for each view with different confidences.

\item Latent Multi-view Subspace Clustering (LMSC)~\cite{8502831} clusters data points with latent representation and simultaneously explores underlying complementary information from multiple views.

\item Consistent and Specific Multi-View Subspace Clustering (CSMSC)~\cite{luo2018consistent} is a novel multi-view subspace clustering method, where consistency and specificity are jointly exploited for subspace representation learning.

\item Multi-view Spectral Clustering via Integrating Nonnegative Embedding and Spectral Embedding (NESE)~\cite{HU2020251} inherits the advantages of both graph-based and matrix factorization methods.
\end{enumerate}

\subsection{Experimental settings}
In this work, we just set the cluster number to the true number of classes, which can be estimated by existing algorithms. We concatenate all views of multi-view datasets to form a new single-view dataset for NCut. NCut, CRMSC, AMGL, and NESE construct the affinity by the Gaussian kernel, where we use the constant $1$ as the Gauss kernel parameter.
The parameter $\lambda$ in CRMSC is  chosen from $0.01$ to $0.05$ with a step $0.01.$ We construct the $k$-nearest neighbor graph for MCAN and SWMC. The $k$ is selected from $3$ to $15.$ For the state-of-the-art multi-view subspace clustering methods LMSC and CSMSC, and our proposed JFLMSC, we choose the parameter values from the set $\{0.00001,0.0001, 0.001, 0.01, 0.1, 1, 10, 100\}.$ To guarantee convergence of algorithms, we set the maximum number of iterations to $200.$ Each algorithm is run 20 times, and then we record their average performance.
\subsection{Clustering results}
\begin{table*}[t]
\begin{center}
\caption{Clstering results on the MSRC-v1 dataset.}
\label{table:MSRC}
\begin{tabular}{cccccc}
\hline\noalign{\smallskip}
\diagbox{Methods}{Metrics} & ACC (\%)& NMI (\%)& ARI (\%)& Precision (\%)& F-score (\%)\\
\hline
\noalign{\smallskip}
NCut~\cite{868688}  &  63.81 &  51.00 & 40.81  &  47.16 & 49.14\\
CRMSC~\cite{10.5555/2986459.2986617} & 78.10& 66.77  &  60.59 &  65.12 & 66.15\\
MCAN~\cite{Nie2017MultiViewCA} &  \uline{88.10}&  \uline{77.74}&  \uline{74.25}&  \uline{80.36}& \uline{77.86}\\
AMGL~\cite{10.5555/3060832.3060884} &   80.00&   73.94&  65.44 &  75.04 & 70.15\\
LMSC~\cite{8502831} &   78.57&  69.69 &  61.82 & 66.18  & 67.20\\
SWMC~\cite{10.5555/3172077.3172245} &  78.10& 73.75&  67.23&  78.29& 72.03\\
CSMSC~\cite{luo2018consistent} &  84.76 &  75.02 &  69.96 & 73.39  & 74.17\\
NESE~\cite{HU2020251}  &  82.38 & 73.28  &  68.52 &  71.47 & 72.65\\
JFLMSC  &   \textbf{91.43}&   \textbf{82.30}&   \textbf{81.06}&  \textbf{83.54}& \textbf{83.69}\\
\hline
\end{tabular}
\end{center}
\end{table*}

\begin{table*}[t]
\begin{center}
\caption{Clstering results on the 100leaves dataset.}
\label{table:100leaves}
\begin{tabular}{cccccc}
\hline\noalign{\smallskip}
\diagbox{Methods}{Metrics} & ACC (\%)& NMI (\%)& ARI (\%)& Precision (\%)& F-score (\%)\\
\hline
\noalign{\smallskip}
NCut~\cite{868688}  &   67.19& 83.76&  56.48& 51.61& 56.93\\
CRMSC~\cite{10.5555/2986459.2986617} & 80.06&  92.12 &  75.55 &  71.01 & 76.56\\
MCAN~\cite{Nie2017MultiViewCA} &   89.95&  94.78&  83.30&  87.89& 83.46\\
AMGL~\cite{10.5555/3060832.3060884} &  82.06&  90.64&  60.34&  83.93 & 62.42\\
LMSC~\cite{8502831} &  77.44& 88.33& 65.15&  65.70& 68.60\\
SWMC~\cite{10.5555/3172077.3172245} &   \uline{91.25}&  \uline{95.35}&  78.44&  \uline{92.16}& 78.67\\
CSMSC~\cite{luo2018consistent} &  79.19 & 89.93  &  72.41 & 69.22  & 72.69\\
NESE~\cite{HU2020251}  &  90.69 &  94.25 &  \uline{85.15}&  84.45 & \uline{84.80}\\
JFLMSC  &  \textbf{95.44}& \textbf{97.83}&  \textbf{93.46}& \textbf{92.22}& \textbf{93.53}\\
\hline
\end{tabular}
\end{center}
\end{table*}

\begin{table*}[t]
\begin{center}
\caption{Clstering results on the Out-Scene dataset.}
\label{table:Out-Scene}
\begin{tabular}{cccccc}
\hline\noalign{\smallskip}
\diagbox{Methods}{Metrics} & ACC (\%)& NMI (\%)& ARI (\%)& Precision (\%)& F-score (\%)\\
\hline
\noalign{\smallskip}
NCut~\cite{868688}  & 40.33&  26.78& 19.86& 26.59& 32.17\\
CRMSC~\cite{10.5555/2986459.2986617} &  60.27 &  46.07&  38.61 & 46.52  & 46.38\\
MCAN~\cite{Nie2017MultiViewCA} &  62.17&  52.06&  41.19&  33.00 & 50.07\\
AMGL~\cite{10.5555/3060832.3060884} &  60.49&  54.97&  42.52&  44.28& 51.09\\
LMSC~\cite{8502831} &   69.12&   51.16&  44.77&   50.62& 51.97\\
SWMC~\cite{10.5555/3172077.3172245} &  60.83& 53.48& 42.65& 50.32& 51.67\\
CSMSC~\cite{luo2018consistent} &  \textbf{73.48} &  \uline{57.96}& \uline{52.19}&  \uline{56.51}& \uline{58.47}\\
NESE~\cite{HU2020251}  &  71.47 & 54.95  & 45.58  & 52.32  & 54.52\\
JFLMSC  & \uline{72.58}&  \textbf{61.08}&  \textbf{54.95}& \textbf{57.13}& \textbf{61.47}\\
\hline
\end{tabular}
\end{center}
\end{table*}

\begin{table*}[t]
\begin{center}
\caption{Clstering results on the ORL dataset.}
\label{table:ORL}
\begin{tabular}{cccccc}
\hline\noalign{\smallskip}
\diagbox{Methods}{Metrics} & ACC (\%)& NMI (\%)& ARI (\%)& Precision (\%)& F-score (\%)\\
\hline
\noalign{\smallskip}
NCut~\cite{868688}  &  65.75&  79.07& 50.71& 50.06& 51.90\\
CRMSC~\cite{10.5555/2986459.2986617} &  73.78&  86.93&  65.89 & 62.41& 66.65\\
MCAN~\cite{Nie2017MultiViewCA} &  72.75&  83.84&  49.46&  75.89& 50.97\\
AMGL~\cite{10.5555/3060832.3060884} &  74.00&  85.20&  56.95&  48.86& 58.15\\
LMSC~\cite{8502831} &  \uline{83.75}&  \uline{92.93}&   \uline{79.94}&  76.13& \uline{80.42}\\
SWMC~\cite{10.5555/3172077.3172245} &  77.25&   88.22&  66.68& \textbf{80.78}& 67.52\\
CSMSC~\cite{luo2018consistent} &  83.00&  90.05 &  75.02 &  71.25& 75.62\\
NESE~\cite{HU2020251}  &  77.00 &  88.14 &  70.11 &  64.19 & 70.85\\
JFLMSC  & \textbf{84.75} &  \textbf{93.31}&  \textbf{81.39}& \uline{77.38}& \textbf{81.83}\\
\hline
\end{tabular}
\end{center}
\end{table*}

In this subsection, we compare the proposed JFLMSC with the competitors in terms of ACC, NMI, ARI, Precision, and F-score, respectively. Tables~\ref{table:MSRC}-\ref{table:ORL} present the clustering results on the four public data sets, respectively, where the best results are highlighted in bold, and the second-best results are underlined for each dataset. A larger value implies better clustering performance for five metrics. In most cases, the proposed JFLMSC method achieves the best clustering performance in comparison to the other state-of-the-art multi-view clustering methods. Also, we have other important findings as follows.
\begin{enumerate}
\item Compared to the single view spectral clustering method NCut, all multi-view clustering methods gain better results. For example, CRMSC outperforms NCut nearly 20\% in terms of ARI on the MSRC-v1 data set. Especially, the improvement of our proposed JFLMSC with respect to NCut is about 28\%, 31\%, 41\%, 36\%, and 34\% on ACC, NMI, ARI, Precision, and F-score, respectively. This phenomenon demonstrates that multi-view clustering methods can better mine the latent clustering structure of the dataset while concatenating all views of multi-view datasets can not effectively leverage information from multiple data sources.

\item Compared to multi-view spectral clustering methods, i.e., CRMSC, AMGL, and NESE, our proposed JFLMSC also shows the best results on four databases. For example, on the MSRC-v1 dataset, the proposed method outperforms state-of-the-art NESE more than 10\% improvement in terms of ACC, ARI, Precision, and F-score. This validates the effectiveness of the proposed method, which can learn a higher quality similarity graph via exploring the global and local structure of data simultaneously.

\item The state-of-the-art multi-view subspace clustering methods LMSC and CSMSC show better performance on ORL and Out-Scene datasets compared to non-subspace clustering methods. For example, CSMSC shows the best result in terms of ACC on the Out-Scene dataset. However, LMSC shows lower performance on the MSRC-v1 dataset compared to the other multi-view clustering methods. CSMSC is laggard behind JFLMSC on the 100leaves dataset. This is because LMSC and CSMSC completely ignore the local structure of data.

\item Our proposed JFLMSC consistently outperforms all competitors in terms of NMI, ARI, and F-score on four datasets. This fact indicates that simultaneously unifying the global representation learning and local structure learning enables JFLMSC to learn the optimal graph for clustering. Specifically, JFLMSC learns weights for different features, which can further improve clustering performance.
\end{enumerate}

\subsection{Ablation study of the proposed method}
\begin{table*}[htbp]
\begin{center}
\caption{Ablation study of the proposed method.}
\label{table:Ablation}
\begin{tabular}{cccccc}
\hline\noalign{\smallskip}
\diagbox{Methods}{Metrics} & ACC (\%)& NMI (\%)& ARI (\%)& Precision (\%)& F-score (\%)\\
\hline
       &\multicolumn{5}{c}{MSRC-v1}  \\
\hline\noalign{\smallskip}
Eq.~\eqref{06}  &  80.95&  73.67&  69.08& 70.29& 73.64\\
Eq.~\eqref{07}  & 83.26&  76.91&  70.76& 70.64& 75.18\\
JFLMSC  &   \textbf{91.43}&   \textbf{82.30}&   \textbf{81.06}&  \textbf{83.54}& \textbf{83.69}\\
\hline
 &\multicolumn{5}{c}{100leaves}  \\
\hline\noalign{\smallskip}
Eq.~\eqref{06}  &  85.88&  92.87&  80.37& 76.47& 80.57\\
Eq.~\eqref{07}  &  94.00&  97.33&  92.04& 90.56& 92.11\\
JFLMSC  &  \textbf{95.44}& \textbf{97.83}&  \textbf{93.46}& \textbf{92.22}& \textbf{93.53}\\
\hline
 &\multicolumn{5}{c}{Out-Scene}  \\
\hline\noalign{\smallskip}
Eq.~\eqref{06}  &  69.53&  56.77&  48.11& 54.22& 54.84\\
Eq.~\eqref{07}  &  72.32&  61.07&  54.71& 56.55& 61.27\\
JFLMSC  &  \textbf{72.58}&  \textbf{61.08}&  \textbf{54.95}& \textbf{57.13}& \textbf{61.47}\\
\hline
 &\multicolumn{5}{c}{ORL}  \\
\hline\noalign{\smallskip}
Eq.~\eqref{06}  &  79.75&  89.02&  72.73& 68.77& 73.39\\
Eq.~\eqref{07}  &  84.50&  91.79&  78.20& 74.89& 78.72\\
JFLMSC  & \textbf{84.75} &  \textbf{93.31}&  \textbf{81.39}& \textbf{77.38}& \textbf{81.83}\\
\hline
\end{tabular}
\end{center}
\end{table*}

In this subsection, we present some ablation study of the proposed method. Specifically, the results only capturing local structure, and the results without considering featurewise weight learning have been analyzed, respectively. Clustering results on the four datasets are shown in Table~\ref{table:Ablation}.

We use Eq.~\eqref{06} to adaptively learn the local structure of each view, while Eq.~\eqref{07} simultaneously learns the global and local structure of each view.
Table~\ref{table:Ablation} demonstrates the improvement of Eq.~\eqref{07} is significant compared to Eq.~\eqref{06}. The clustering ACC of Eq~\eqref{07} on 100leaves dataset achieves more than 9\% in comparison with Eq.~\eqref{06}. Similarly, Eq~\eqref{07} outperforms Eq~\eqref{06} nearly 7\% on the Outdoor Scene data set in terms of F-score.

Furthermore, we can see that our proposed JFLMSC achieves the best clustering results. In particular, on the MSRC-v1 data set, the proposed method achieves more than 10\% of ARI and Precision in comparison with the second-best. In conclusion, the proposed method effectively captures the global and local structure of multi-view data, and if the original data have noisy, JFLMSC can adaptively assign small weights for noisy features and large weights for discriminative features. Thus JFLMSC outperforms Eq.~\eqref{06} and Eq.~\eqref{07}.

\subsection{Parameter and convergence analysis}

\begin{figure*}[t!]
\centerline{\includegraphics[scale=0.7]{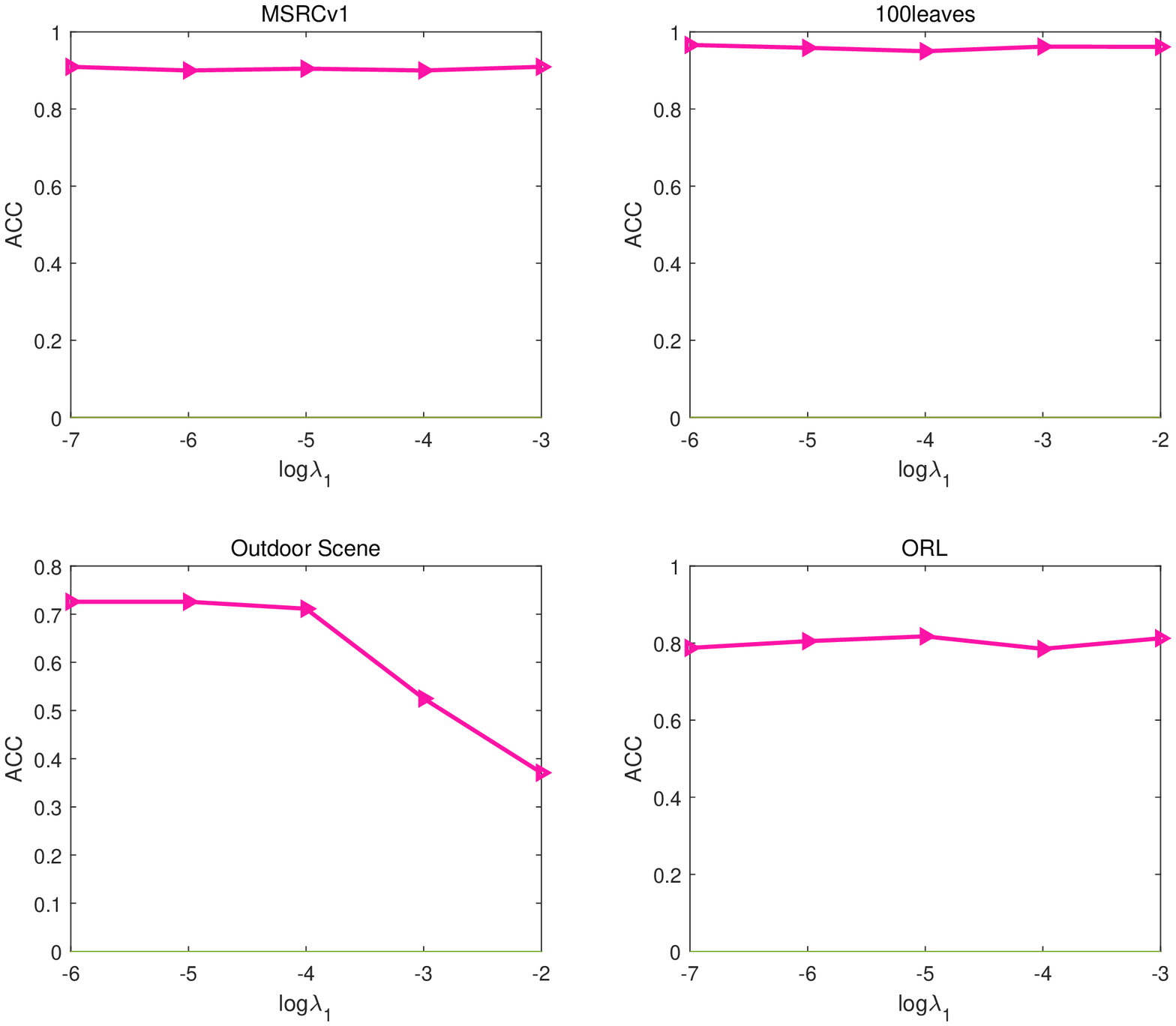}}
\caption{Clustering ACC versus different values of parameter $\lambda_1$.}
\label{fig:lambda1}
\end{figure*}

\begin{figure*}[htbp]
\centerline{\includegraphics[scale=0.7]{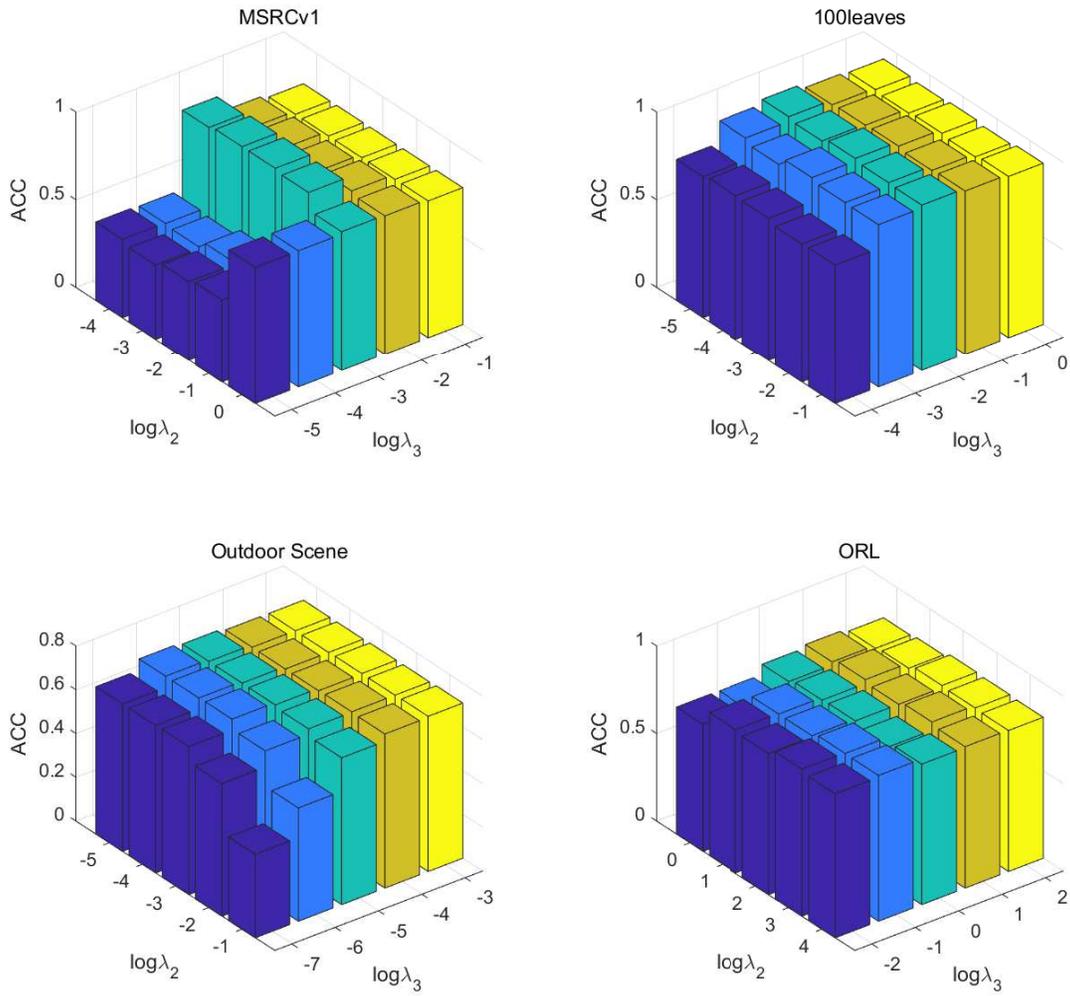}}
\caption{Clustering ACC versus different values of parameters $\lambda_2$ and $\lambda_3$.}
\label{fig:lambda23}
\end{figure*}

\begin{figure*}[htbp]
\centerline{\includegraphics[scale=0.7]{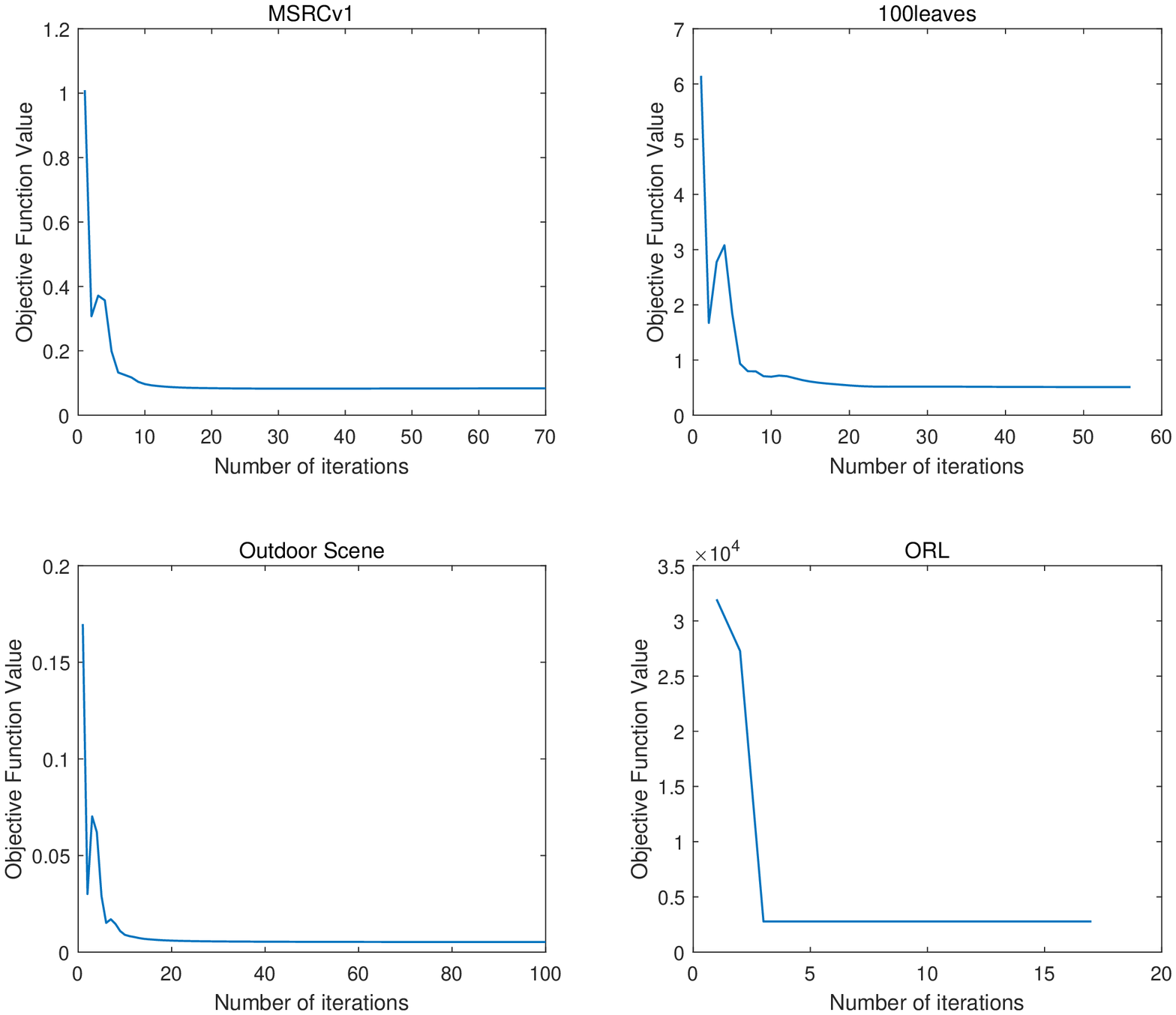}}
\caption{Convergence analysis of our proposed method.}
\label{fig:COV}
\end{figure*}

For the proposed JFLMSC, there are three parameters in Eq.~\eqref{08} to balance the importance of $2$-norm constraint term, error term, and the consistency term of
all views. In this section, we investigate how the clustering ACC varies with the change of these three parameters. In order to deeply study the influence of parameters on clustering performance, we select parameter values from $\{1e-7,1e-6,1e-5,\ldots,1e-1,1\}.$

We first fix parameters $\lambda_2$ and $\lambda_3$, and then perform the proposed JFLMSC with different values of $\lambda_1$ to observe the influence of $\lambda_1$ to the clustering
ACC. From Fig.~\ref{fig:lambda1}, we can see that ACC is insensitive to $\lambda_1$ when $\lambda_1\leq 1e-4.$ Attentively, when $\lambda_1> 1e-4,$ ACC drops sharply on the Outdoor Scene dataset. This is because the consistency term may destroy diversity learning of each view if $\lambda_1$ is too large.

Fig.~\ref{fig:lambda23} show ACC versus different values of $\lambda_2$ and $\lambda_3$ with fixed $\lambda_1$. It is observed that ACC of the proposed JFLMSC is sensitive with $\lambda_2$ and $\lambda_3.$ Different datasets need different parameter combinations to achieve the best performance. This indicates that selecting $\lambda_2$ and $\lambda_3$ is data-driven.

Fig.~\ref{fig:COV} exhibits the trend of the objective value computed by the proposed Algorithm~\ref{algo} with respect to the number of iterations on four datasets, respectively.
The result empirically confirms the convergence behavior of Algorithm~\ref{algo}.

\section{Conclusion}
In this paper, we presented a novel multi-view subspace clustering method named Joint Featurewise Weighting and Lobal Structure Learning for Multi-view Subspace Clustering (JFLMSC).
Specifically, adaptive local learning, self-representation learning, weight learning for features are seamlessly integrated into a unified framework to exploit the global and local structure of each view for multi-view subspace clustering. In the proposed method, discriminative features are assigned high weights, while noisy features have low weights. As a result, we can obtain a high-quality similarity graph for each view, which also extracts the consistency information across different views using the consensus clustering assignment matrix. Extensive experiment analysis is carried out on four benchmark multi-view datasets in terms of performance comparison, ablation study, sensitivity analysis, and convergence analysis, which demonstrate the superiority of our proposed method.

\bibliographystyle{elsarticle-num}
\bibliography{JSCLMC}
\end{document}